\documentclass{article} % For LaTeX2e
\usepackage{iclr2026_conference,times}

\usepackage{amsmath,amsfonts,bm}

\def\eqref#1{equation~\ref{#1}}
\def\1{\bm{1}}

\DeclareMathAlphabet{\mathsfit}{\encodingdefault}{\sfdefault}{m}{sl}
\SetMathAlphabet{\mathsfit}{bold}{\encodingdefault}{\sfdefault}{bx}{n}

\usepackage{hyperref}
\usepackage{url}
\usepackage{multirow}
\usepackage{booktabs}
\usepackage[table]{xcolor}
\usepackage{tabularx}
\usepackage{diagbox}
\usepackage{wrapfig}
\usepackage{enumitem}
\usepackage{graphicx}

\title{GraphSearch: An Agentic Deep Searching Workflow for Graph Retrieval-Augmented Generation}

\author{
 \textbf{Cehao Yang\textsuperscript{1,2,3}\footnotemark[1]},
 \textbf{ Xiaojun Wu\textsuperscript{1,2,3}\footnotemark[1] },
 \textbf{ Xueyuan Lin\textsuperscript{1,2,4}\footnotemark[1]},
 \textbf{ Chengjin Xu\textsuperscript{1,3}}\footnotemark[2],
 \textbf{ Xuhui Jiang\textsuperscript{1,3}},\\
 \textbf{ Yuanliang Sun\textsuperscript{3}}
 \textbf{ Jia Li\textsuperscript{2}},
 \textbf{ Hui Xiong\textsuperscript{2}}\footnotemark[2],
 \textbf{ Jian Guo\textsuperscript{1}}\footnotemark[2]
\\
 \textsuperscript{1}IDEA Research, International Digital Economy Academy
 \\
 \textsuperscript{2}Hong Kong University of Science and Technology (Guangzhou)
\\
 \textsuperscript{3}DataArc Tech Ltd.
 \\
 \textsuperscript{4}Hithink RoyalFlush Information Network Co., Ltd
 \\
 \texttt{\{cyang289,xwu647,xlin058,jialee\}@connect.hkust-gz.edu.cn},
 \\
 \texttt{xionghui@ust.hk}, \texttt{\{xuchengjin,jiangxuhui,guojian\}@idea.edu.cn}
}
\iclrfinalcopy % Uncomment for camera-ready version, but NOT for submission.
\begin{document}

\maketitle
\footnotetext[1]{Equal contribution.}
\footnotetext[2]{Corresponding authors.}
\footnotetext[3]{Our code implementation are available at \href{https://github.com/DataArcTech/GraphSearch}{https://github.com/DataArcTech/GraphSearch}.}

\begin{abstract}
Graph Retrieval-Augmented Generation (GraphRAG) enhances factual reasoning in LLMs by structurally modeling knowledge through graph-based representations. However, existing GraphRAG approaches face two core limitations: shallow retrieval that fails to surface all critical evidence, and inefficient utilization of pre-constructed structural graph data, which hinders effective reasoning from complex queries. To address these challenges, we propose \textsc{GraphSearch}, a novel agentic deep searching workflow with dual-channel retrieval for GraphRAG. \textsc{GraphSearch} organizes the retrieval process into a modular framework comprising six modules, enabling multi-turn interactions and iterative reasoning. Furthermore, \textsc{GraphSearch} adopts a dual-channel retrieval strategy that issues semantic queries over chunk-based text data and relational queries over structural graph data, enabling comprehensive utilization of both modalities and their complementary strengths. Experimental results across six multi-hop RAG benchmarks demonstrate that \textsc{GraphSearch} consistently improves answer accuracy and generation quality over the traditional strategy, confirming \textsc{GraphSearch} as a promising direction for advancing graph retrieval-augmented generation.
\end{abstract}
\section{Introduction}
\begin{wrapfigure}{r}{0.4\textwidth}
\vspace{-12pt}
\centering
\includegraphics[width=0.4\textwidth]{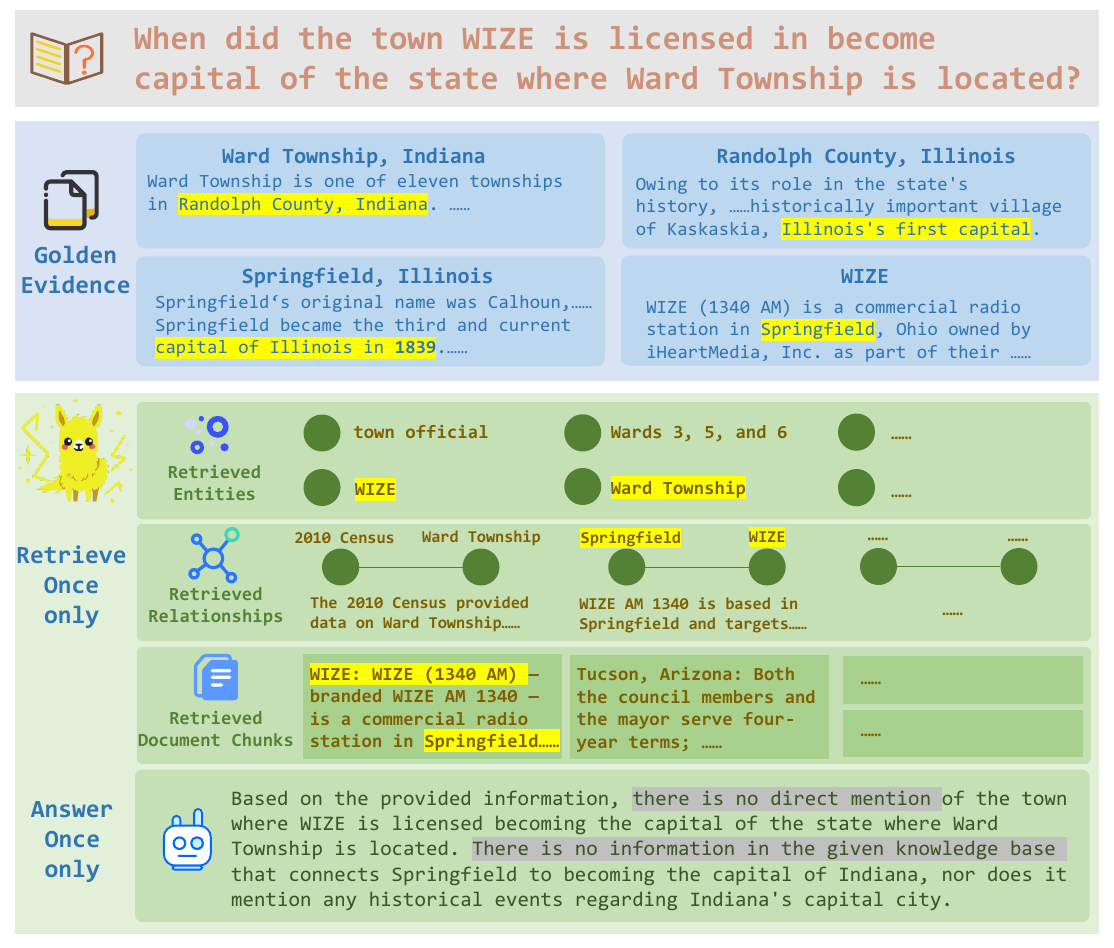}
\vspace{-13pt}
\caption{Shallow retrieval.}
\label{fig:challenge}
\vspace{-15pt}
\end{wrapfigure}
Large Language Models (LLMs) demonstrates remarkable capabilities in natural language understanding and reasoning~\citep{llm1,llm2}. Despite their strong performance, LLMs inherently rely on their parametric knowledge, which often results in hallucinations and a lack of factual grounding~\citep{hallucination,factuality}. Retrieval-augmented generation (RAG) has emerged as a paradigm that combines LLMs with external knowledge bases, enhancing factuality, credibility and interpretability in knowledge-intensive tasks~\citep{rag}.

More recently, Graph Retrieval-Augmented Generation (GraphRAG) is introduced to overcome the shortcomings of traditional RAG, which relies solely on semantic similarity for retrieval~\citep{graphragsurvey}. By constructing structural graph knowledge bases (graph KBs) and leveraging hierarchical retrieval strategies, GraphRAG strengthens the integration of contextual information across massive entities and relationships~\citep{raptor,graphrag,lightrag}. Building upon this foundation, some advanced graph-based enhancements that incorporate diverse structures, including heterogeneous graphs, causal graphs, and hypergraphs, to enrich representational ability and facilitate more abundant graph construction~\citep{minirag,causalrag,hypergraphrag,hyperrag,noderag}. In addition, heuristic strategies such as path-based search, pruning, and memory-inspired indexing further reinforce reasoning abilities and enable deeper multi-step exploration~\citep{pathrag,hipporag,hipporag2,proprag}.

\begin{figure}[t]
\centering
\includegraphics[width=\linewidth]{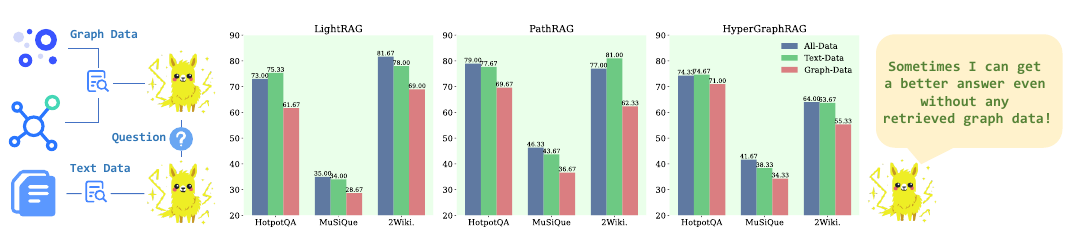}
\caption{Comparison of using graph data only, text data only, or all data as commonly adopted in GraphRAG approaches. The metric is SubEM. The contribution of retrieved graph data is marginal.}
\label{fig:rethinking}
\vspace{-10pt}
\end{figure}

However, existing GraphRAG approaches still face challenges that lead to performance bottlenecks: (i) \textbf{Shallow retrieval results in missing evidence for complex queries.} Most GraphRAG methods adopt a single-round retrieval-and-generation process as the interaction strategy between the LLM and the graph KB~\citep{graphrag, lightrag, minirag}. However, as illustrated in Figure~\ref{fig:challenge}, when handling a complex query that requires four pieces of golden evidence, \textit{``When did the town WIZE is licensed in become capital of the state where Ward Township is located?''}, the entity \textit{Randolph County} is not retrieved by the LightRAG retriever. Consequently, the LLM’s reasoning suffers from broken logic and insufficient evidence. (ii) \textbf{Limited ability to exploit structural data due to constrained retrieval scope}. Existing GraphRAG methods with heuristic path-construction schemes~\citep{minirag, pathrag, hipporag} often fail to fully leverage the structural information in graph KBs, fundamentally because shallow retrieval restricts the coverage of relevant nodes and relations. Without sufficient coverage of retrieved subgraphs, the available structural signals are fragmented and sparse, making it difficult for LLMs to integrate semantic and structural modalities for complex reasoning. As shown in Figure~\ref{fig:rethinking}, models may perform comparably with text-only evidence, highlighting that the underutilization of graph data is tightly coupled with the limitations of current retrieval strategies.

We propose \textbf{\textsc{GraphSearch}, an agentic deep searching workflow for GraphRAG}. As illustrated in Figure~\ref{fig:workflow}, \textsc{GraphSearch} is a novel agent framework designed to access graph KBs through dual-channel retrieval, acquiring both semantic and structural information, and performing multi-turn interactions to complete complex reasoning tasks. Targeting the shallow retrieval problem in existing GraphRAG approaches, \textbf{\textsc{GraphSearch} models retrieval as a modular searching pipeline}, which consists of six modules: \textit{Query Decomposition (QD)}, \textit{Context Refinement (CR)}, \textit{Query Grounding (QG)}, \textit{Logic Drafting (LD)}, \textit{Evidence Verification (EV)}, and \textit{Query Expansion (QE)}. Through the coordinated contributions of these modules, \textsc{GraphSearch} decomposes complex queries into tractable atomic sub-queries, retrieves fine-grained knowledge from graph KBs, and iteratively performs logical reasoning and reflection to remedy missing evidence. Furthermore, \textbf{\textsc{GraphSearch} adopts a dual-channel retrieval strategy}, constructing semantic queries over chunk-based text data and relational queries over structural graph data, thereby fully synergizing both modalities and integrating them into contexts that support LLMs in complex reasoning.

We conduct experiments on six multi-hop RAG datasets. The results demonstrate that leveraging the graph KBs retrievers built upon the corresponding GraphRAG approaches, \textsc{GraphSearch} consistently outperforms the single-round interaction strategy in terms of answer accuracy and generation quality, while also exhibiting strong plug-and-play capability, as shown in Table~\ref{Tab:main_exp}. Furthermore, the effectiveness of the dual-channel retrieval strategy, the contributions of agentic modules, and its robustness under a small-scale LLM and varying retrieval budgets are all empirically validated.

% Our main contributions are summarized as follows:  
% \begin{itemize}[leftmargin=*]
%     \item We propose \textsc{GraphSearch}, an agentic deep searching workflow that overcomes the challenges of shallow retrieval and the ineffective use of graph data in existing GraphRAG approaches.

%     \item We introduce a modular searching pipeline with coordinated modules for iterative reasoning and a dual-channel retrieval strategy integrating semantic and relational queries to exploit graph KBs.  

%     \item Experiment results across six multi-hop RAG datasets demonstrating that \textsc{GraphSearch} consistently outperforms vanilla GraphRAG baselines in answer accuracy and generation quality.
% \end{itemize}

% Our contributions are as follows: (i) We propose \textsc{GraphSearch}, an agentic deep searching workflow that overcomes the challenges of shallow retrieval and the ineffective use of graph data in existing GraphRAG approaches. (ii) We introduce a modular searching pipeline with coordinated modules for iterative reasoning and a dual-channel retrieval strategy integrating semantic and relational queries over graph KBs. (iii) Experiment results across six multi-hop RAG datasets demonstrating that \textsc{GraphSearch} consistently outperforms vanilla GraphRAG in accuracy and quality.
\section{Related Work}
\subsection{Graph Retrieval-Augmented Generation}
RAG augments LLMs with external evidence to improve factuality of knowledge-intensive tasks~\citep{rag}. Building on this, GraphRAG is an advance paradigm that explicitly models structural relations among entities, thereby capturing relational semantics, contextual dependencies and structural knowledge integration~\citep{graphragsurvey,graphrag}.
Early work~\citep{raptor, graphrag} emphasize hierarchical summarization and global information integration, but they insufficiently leveraged the fine-grained structural information. LightRAG~\citep{lightrag} advanced this direction by incorporating graph structures into both indexing and retrieval.
Recent efforts in graph KB construction introduce diverse structural representations, such as the design of heterogeneous and lightweight graph structures~\citep{minirag, noderag}, the extension to hypergraphs that capturing higher-order relational dependencies~\citep{hypergraphrag, hyperrag}, and the leverage of causal graphs to improve logical continuity~\citep{causalrag}. Additionally, retrieval strategies on graph KBs increasingly rely on heuristic path exploration, such as the topology-enhanced lightweight search~\citep{minirag}, the pruning via relational path retrieval~\citep{pathrag}, the utilization of personalized memory-inspired reasoning~\citep{hipporag, hipporag2}, and the adoption of beam search over proposition paths~\citep{proprag}.
Despite these advances, current GraphRAG approaches remain constrained by shallow retrieval, limiting their ability to perform deep searching over graph KBs.

\subsection{Agentic Retrieval-Augmented Generation}

RAG improves factual grounding by retrieving external knowledge~\citep{rag}, but single-round interaction is insufficient for complex reasoning tasks. Early advances focus on atomic-level improvements of RAG in query decomposition~\citep{decomposition}, query rewriting~\citep{queryrewrite,rqrag}, retrieval compression~\citep{recomp}, and selective retrieval decisions~\citep{slimplm}, which refine the retrieval process at a fine granularity. Beyond these, modular RAG systems~\citep{modularrag,flashrag,composerag} have been proposed to flexibly reconfigure retrieval and reasoning modules into composable pipelines. More recently, agentic approaches emerged, enabling LLMs to iteratively plan, retrieve, and reflect. Representative methods include reasoning–acting synergy in ReAct~\citep{react}, self-reflective retrieval in Self-RAG~\citep{selfrag}, test-time planning in PlanRAG~\citep{planrag}, and reinforcement-learned search agents in Search-o1~\citep{searcho1} and Search-r1~\citep{searchr1}. Subsequently, pioneering works~\citep{tog, tog2, gear, hybgrag} integrated structural graph knowledge for retrieval into the agentic RAG workflow to support the multi-step reasoning.

\section{Preliminaries}

\paragraph{Graph Knowledge Database.}  
Given a document collection $D$, the graph indexer $\phi$ segments $D$ into a set of text chunks $K$. For each chunk $k \in K$, an extractor $\mathcal{R} \in \phi$ identifies a set of entities $e = \{ e_{\text{name}}, e_{\text{prop}}, e_{\text{desc}} \}$. For any pair of entities $e_h, e_t \in k$, a relation is defined as $r = \{ e_h, e_t, r_{\text{prop}}, r_{\text{desc}} \}$. Aggregating all entities and relations yields the graph KB $G = \{E, R, K\}$, where $E$ denotes the entity set, $R$ the relation set, and $K$ the associated chunk-level textual context.

\paragraph{Graph KB Retrieval.}
Given a query $q$, a graph KB retriever $\psi$ selects a relevant context set $C = \{E_q, R_q, K_q\} \subset G$ that maximizes semantic relevance to $q$. The retriever aims to return structural graph data and chunk-based text data that provide sufficient evidence for answer generation.

\paragraph{LLM Answer Generation.}
The language model consumes the query $q$ together with the retrieved context $C$ to generate an output $y$. The generation is modeled as $P(y \mid q) = \sum_{C \subset G} P(y \mid q, C)\, P(C \mid q, G)$, where $P(C \mid q, G)$ represents the retrieval probability over the graph KB, and $P(y \mid q, C)$ denotes the generation probability conditioned on the integrated evidence.

\begin{figure}[t]
\centering
\includegraphics[width=\linewidth]{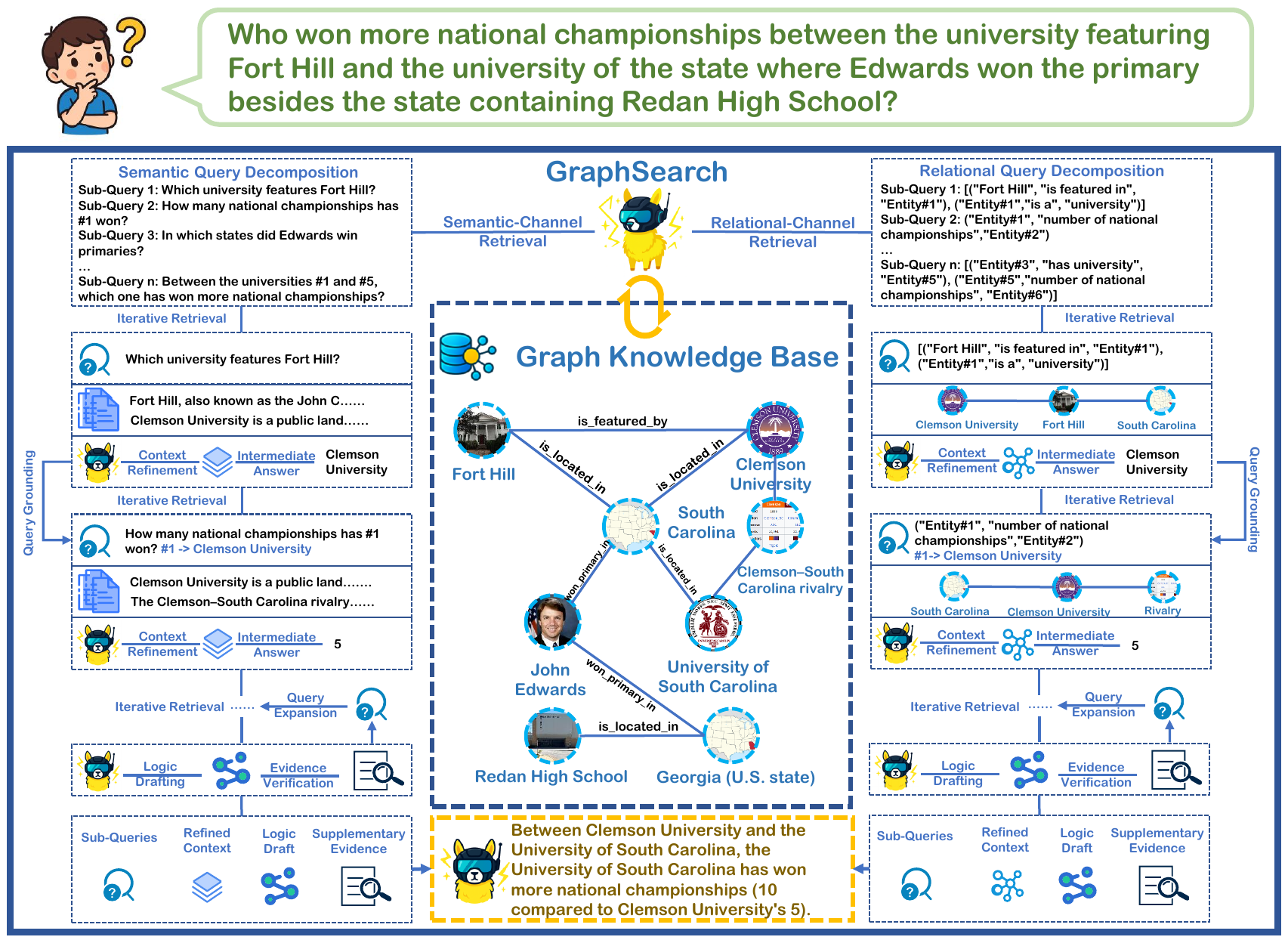}
\caption{Overview of our \textsc{GraphSearch} framework.}
\label{fig:workflow}
\vspace{-10pt}
\end{figure}

\section{GraphSearch}
 
The overview of \textsc{GraphSearch} is shown in Figure~\ref{fig:workflow}. We build upon existing GraphRAG methods to construct the graph KB from documents. On top of this, \textsc{GraphSearch} leverages the GraphRAG retriever to perform deep searching, thereby enabling better answer generation.

\subsection{The Modular Deep Searching Pipeline}
\subsubsection{Iterative Retrieval}

\paragraph{Query Decomposition.}
Given a complex query $Q$ as input, the goal of this module is to decompose $Q$ into a sequence of atomic sub-queries $\{q_1, q_2, \dots, q_m\} = \mathrm{P}_{\mathrm{QD}}(Q)$ prompted by template $\mathrm{P}_{\mathrm{QD}}$, each representing a smaller and tractable component of the original question. In practice, each $q_i$ focuses on resolving a single entity, relation, or contextual dependency, thereby enabling the retriever to access fine-grained evidence and reducing the reasoning complexity of the overall task. For each sub-query $q_i$, the graph KB retriever $\psi$ accesses database $G$ to return 
\begin{equation}
    C_{q_i} = \psi(q_i \mid G) = \{E_{q_i}, R_{q_i}, K_{q_i}\}
\end{equation}
where $C_{q_i}$ is the retrieved context of sub-query $q_i$. The detail of prompt $\mathrm{P}_{\mathrm{QD}}$ is in Figure~\ref{fig:prompts_1}.

\paragraph{Context Refinement.}
Once the initial context $C_{q_i}$ is retrieved for a sub-query $q_i$, this module aims to refine the evidence by filtering redundant information and highlighting the most relevant entities, relations, and textual chunks. Given that raw retrieval, the refined context is obtained as $C_{q_i}' = \mathrm{P}_{\mathrm{CR}}(q_i, C_{q_i})$. This operation ensures that each refined context $C_{q_i}'$ contains only the most informative evidence for answering, thereby improving grounding quality in subsequent reasoning.

\paragraph{Query Grounding.}
The sub-queries $\{q_1, q_2, \dots, q_m\}$ are designed to be semantically independent yet logically ordered, such that the answer to one sub-query can serve as contextual grounding for subsequent ones. In practice, many decomposed queries may contain placeholders or unresolved references that depend on the answers of prior sub-queries. To resolve this, each $q_i$ is first paired with its retrieved context $C_{q_i}$ and produce an intermediate answer $\hat{a}_{q_i} = \mathrm{LLM}(q_i, C_{q_i})$, then progressively accumulated to support later queries. Formally, the grounded query is expressed as
\begin{equation}
    \tilde{q}_i = \mathrm{P}_{\mathrm{QG}}(q_i, \{q_{<i}, C_{q_{<i}}, \hat{a}_{q_{<i}}\}),
\end{equation}
This procedure guarantees that each $\tilde{q}_i$ is contextually instantiated rather than under-specified, enabling the graph KB retriever to fetch a more relevant context $C_{\tilde{q}_i}$ for subsequent reasoning.

\subsubsection{Reflection Routing}

\paragraph{Logic Drafting.}
The role of this module is to organize these pieces into a coherent reasoning chain that outlines how partial answers connect to the original query $Q$. Specifically, the drafting prompt $\mathrm{P}_{\mathrm{LD}}$ integrates the sequence of \{$q_i, C_{\tilde{q}_i}, \hat{a}_{q_i}$\} to produce a structured draft $\mathcal{L}$, where
\begin{equation}
    \mathcal{L} = \mathrm{P}_{\mathrm{LD}}(\{\tilde{q}_i, C_{\tilde{q}_i}, \hat{a}_{q_i}\}_{i=1}^m).
\end{equation}
During this drafting process, the module not only consolidates available evidence but also exposes potential gaps in the reasoning chain. For instance, if a sub-query relies on entities or relations that were not retrieved in earlier steps, or if the accumulated sub-queries with intermediate answers $\{\tilde{q}_i, \hat{a}_{q_i}\}$ form an inconsistent chain, such deficiencies are explicitly reflected in $\mathcal{L}$ and exposed.

\paragraph{Evidence Verification.}
This module evaluates whether the accumulated evidence in $\mathcal{L}$ is sufficient and logically consistent to support a final answer. The verification prompt $\mathrm{P}_{\mathrm{EV}}$ inspects both the retrieved contexts and the intermediate answers, checking for factual grounding, coherence, and potential contradictions. Formally, this process can be described as
\begin{equation}
    \mathcal{V} = \mathrm{P}_{\mathrm{EV}}(\{\tilde{q}_i, C_{\tilde{q}_i}, \hat{a}_{q_i}\}_{i=1}^m, \mathcal{L}),
\end{equation}
where $\mathcal{V} \in \{\mathrm{Accept}, \mathrm{Reject}\}$ denotes the verification decision, the former implying that the reasoning chain is logically reliable, and the latter indicating missing or inconsistent evidence.

\paragraph{Query Expansion.}
This module generates additional sub-queries that explicitly target the missing evidence. Formally, using the expansion prompt and outputs a set of expanded sub-queries
\begin{equation}
    \{{q_j^{+}}\}_{j=1}^n = \mathrm{P}_{\mathrm{QE}}(\{\tilde{q}_i, C_{\tilde{q}_i}, \hat{a}_{q_i}\}_{i=1}^m, \mathcal{L}).
\end{equation}
Each expanded sub-query $q_j^{+}$ is submitted to the retriever $\psi$, yielding supplementary evidence $C_{q_i^{+}} = \psi(q_i^{+} \mid G) = \{E_{q_i^{+}}, R_{q_i^{+}}, K_{q_i^{+}}\}$. The additional contexts $C_{q_i^{+}}$ are appended, thereby enriching the evidence pool and ensuring that knowledge gaps revealed in $\mathcal{L}$ can be actively filled, leading to a more reliable reasoning process.

\subsection{Dual-Channel Retrieval}

\paragraph{Semantic Queries.}  
The semantic channel emphasizes retrieving descriptive evidence from chunk-level text. Given a complex query such as \textit{``How many times did plague occur in the place where the creator of \textit{The Worship of Venus} died?''}, the retriever first reformulates it into a sequence of semantically coherent sub-queries $\{q_1^{(s)}, q_2^{(s)}, \dots, q_m^{(s)}\}$. Each $q_i^{(s)}$ is resolved against the text corpus as $C_{q_i^{(s)}} = \{K_{q_i^{(s)}}\}$, focusing on a single factual aspect, such as identifying the creator of the artwork, locating the place where this creator died, and finally retrieving records about the frequency of plague occurrences in that place. This design allows the semantic channel to capture nuanced descriptive information scattered across the corpus, ensuring that the retrieved textual evidence provides sufficient coverage for each step of reasoning.

\paragraph{Relational Queries.}  
The relational channel formulates the same problem directly in terms of structured triples. Given a complex query $Q$, it is decomposed into a sequence of relational sub-queries $\{q_1^{(r)}, q_2^{(r)}, \dots, q_n^{(r)}\}$, each mapped into subject--predicate--object relations. For each $q_j^{(r)}$, the retriever returns a subgraph context $C_{q_j^{(r)}} = \{E_{q_j^{(r)}}, R_{q_j^{(r)}}\}$, focusing only on entities and relations. For example, the painting \textit{The Worship of Venus} $\rightarrow$ its creator $\rightarrow$ place of death $\rightarrow$ plague occurrences. Unresolved references (e.g., $\mathrm{Entity\#1}, \mathrm{Entity\#2}$) are progressively instantiated once upstream triples are resolved. This explicit traversal enforces logical dependencies and supports multi-hop reasoning, enabling the retriever to surface subgraphs that directly encode the answer path with reduced reliance on textual co-occurrence.

\section{Experiments}
\subsection{Experimental Setup}
\paragraph{Datasets.} 
To evaluate the performance of \textsc{GraphSearch}, we conducted experiments on six multi-hop QA benchmarks within the RAG setting. The \textbf{Wikipedia}-based benchmarks include \textbf{HotpotQA}~\citep{hotpotqa}, \textbf{MuSiQue}~\citep{musique}, and \textbf{2WikiMultiHopQA}~\citep{2wiki} following~\citep{hipporag2, longfaith}. The \textbf{Domain}-based benchmarks~\citep{ultradomain} incorporate multi-hop questions synthesized by~\citep{hypergraphrag}, covering fields like \textbf{Medical}, \textbf{Agriculture}, and \textbf{Legal}. More details are provided in the Appendix~\ref{app:datasets}.

\begin{table*}[t]
\caption{Experiment results across six multi-hop QA benchmarks covering \textbf{Wikipedia}-based and \textbf{Domain}-based datasets. The + means \textbf{\textsc{GraphSearch}} integrates with various graph KB retrievers built upon the corresponding GraphRAG methods. The backbone LLM is \textit{Qwen2.5-32B-Instruct}.}
\label{Tab:main_exp}
\setlength{\tabcolsep}{4pt}
\begin{center}
% \small
\begin{tabular}{lccccccccc}
\toprule
\multirow{2}{*}{\normalsize\textbf{Method}}
& \multicolumn{3}{c}{\textbf{HotpotQA}}
& \multicolumn{3}{c}{\textbf{MuSiQue}}
& \multicolumn{3}{c}{\textbf{2WikiMultiHopQA}} \\
\cmidrule[1pt](lr){2-4} \cmidrule[1pt](lr){5-7} \cmidrule[1pt](lr){8-10}
& {\scriptsize $\mathrm{SubEM}$} & {\scriptsize $\mathrm{A\text{-}Score}$} & {\scriptsize $\mathrm{E\text{-}Score}$}
& {\scriptsize $\mathrm{SubEM}$} & {\scriptsize $\mathrm{A\text{-}Score}$} & {\scriptsize $\mathrm{E\text{-}Score}$}
& {\scriptsize $\mathrm{SubEM}$} & {\scriptsize $\mathrm{A\text{-}Score}$} & {\scriptsize $\mathrm{E\text{-}Score}$} \\
\midrule
Vanilla LLM      & 33.67 & 6.90 & 5.98 & 12.33 & 6.10 & 5.87 & 48.33 & 6.95 & 4.50 \\
Naive RAG        & 72.00 & 8.88 & 9.04 & 40.00 & 7.21 & 8.18 & 72.33 & 7.93 & 8.03 \\
ReAct            & 33.33 & -    & -    & 16.00 & -    & -    & 51.33  & -    & -    \\
\midrule
\multicolumn{10}{c}{\textbf{GraphRAG Baselines}} \\
\midrule
GraphRAG         & 72.67 & 8.18 & 8.65 & 36.67 & 6.58 & 7.32 & 79.33 & 7.44 & 7.99 \\
LightRAG         & 73.00 & 8.30 & 8.66 & 35.00 & 6.50 & 7.28 & 81.67 & 7.62 & 7.94 \\
MiniRAG          & 68.00 & 7.95 & 8.24 & 41.00 & 6.93 & 7.67 & 74.00 & 7.57 & 7.61 \\
PathRAG          & 79.00 & 8.99 & 9.17 & 46.33 & 7.26 & 8.02 & 77.00 & 8.25 & 8.34 \\
HippoRAG2        & 76.67 & 8.45 & 8.73 & 44.00 & 7.07 & 7.88 & 72.33 & 7.98 & 8.01 \\
HyperGraphRAG    & 74.33 & 7.39 & 8.69 & 41.67 & 6.76 & 7.53 & 64.00 & 7.62 & 7.80 \\
\midrule
\multicolumn{10}{c}{\textbf{\textsc{GraphSearch}}} \\
\midrule
\rowcolor{gray!20}
+ LightRAG       & 79.00 & 9.21 & \textbf{9.46} & 51.00 & 7.72 & 8.38 & 85.00 & 9.21 & 9.12 \\
\rowcolor{gray!20}
+ PathRAG        & \textbf{82.00} & \textbf{9.24} & 9.42 & \textbf{55.33} & \textbf{7.83} & \textbf{8.48} & \textbf{88.67} & \textbf{9.32} & \textbf{9.29} \\
\rowcolor{gray!20}
+ HyperGraphRAG  & 80.33 & 9.19 & 9.35 & 49.33 & 7.73 & 8.22 & 83.33 & 8.84 & 8.75 \\
\bottomrule
\toprule
\multirow{2}{*}{\normalsize\textbf{Method}}
& \multicolumn{3}{c}{\textbf{Medicine}}
& \multicolumn{3}{c}{\textbf{Agriculture}}
& \multicolumn{3}{c}{\textbf{Legal}} \\
\cmidrule[1pt](lr){2-4} \cmidrule[1pt](lr){5-7} \cmidrule[1pt](lr){8-10}
& {\scriptsize $\mathrm{SubEM}$} & {\scriptsize $\mathrm{A\text{-}Score}$} & {\scriptsize $\mathrm{E\text{-}Score}$}
& {\scriptsize $\mathrm{SubEM}$} & {\scriptsize $\mathrm{A\text{-}Score}$} & {\scriptsize $\mathrm{E\text{-}Score}$}
& {\scriptsize $\mathrm{SubEM}$} & {\scriptsize $\mathrm{A\text{-}Score}$} & {\scriptsize $\mathrm{E\text{-}Score}$} \\
\midrule
Vanilla LLM      & 21.29 & 7.14 & 7.57 & 29.88 & 7.10 & 7.38 & 37.11 & 7.02 & 7.43 \\
Naive RAG        & 54.34 & 8.23 & 8.67 & 54.24 & 7.91 & 8.26 & 53.36 & 7.37 & 7.67 \\
ReAct            & 19.73 & -    & -    & 25.99 & -    & -    & 30.86  & -    & -    \\
\midrule
\multicolumn{10}{c}{\textbf{GraphRAG Baselines}} \\
\midrule
GraphRAG         & 53.32 & 7.59 & 7.98 & 57.81 & 7.84 & 7.66 & 58.98 & 7.57 & 7.23 \\
LightRAG         & 49.80 & 7.36 & 7.57 & 55.66 & 7.38 & 7.32 & 56.84 & 7.01 & 6.78 \\
MiniRAG          & 56.84 & 8.13 & 8.51 & 59.38 & 8.08 & 8.08 & 61.91 & 7.70 & 7.50 \\
PathRAG          & 58.79 & 8.18 & 8.32 & 61.13 & 8.22 & 8.23 & 62.30 & 7.96 & 7.91 \\
HippoRAG2        & 55.08 & 7.90 & 8.03 & 58.20 & 7.95 & 7.86 & 64.45 & 8.02 & 7.81 \\
HyperGraphRAG    & 62.11 & 8.39 & 8.70 & 63.67 & 8.35 & 8.49 & 66.60 & 8.18 & 8.18 \\
\midrule
\multicolumn{10}{c}{\textbf{\textsc{GraphSearch}}} \\
\midrule
\rowcolor{gray!20}
+ LightRAG       & 65.88 & 8.61 & 8.80 & 63.53 & 8.52 & 8.48 & 71.68 & 8.45 & 8.52 \\
\rowcolor{gray!20}
+ PathRAG        & 70.12 & 8.59 & 8.82 & 69.34 & 8.63 & 8.78 & 74.41 & 8.32 & 8.49 \\
\rowcolor{gray!20}
+ HyperGraphRAG  & \textbf{73.24} & \textbf{8.87} & \textbf{9.24} & \textbf{73.83} & \textbf{8.93} & \textbf{9.02} & \textbf{78.52} & \textbf{8.76} & \textbf{8.83} \\
\bottomrule
\end{tabular}
\vspace{-20pt}
\end{center}
\end{table*}

\paragraph{Baselines.} 
We compare \textsc{GraphSearch} with several baseline methods, including \textbf{Vanilla LLM}, \textbf{Naive RAG}~\citep{rag}, \textbf{GraphRAG}~\citep{graphrag}, \textbf{LightRAG}~\citep{lightrag}, \textbf{MiniRAG}~\citep{minirag}, \textbf{PathRAG}~\cite{pathrag}, \textbf{HippoRAG2}~\citep{hipporag2}, and \textbf{HyperGraphRAG}~\citep{hypergraphrag}. More details are provided in the Appendix~\ref{app:baselines}.

\paragraph{Evaluation Metrics.} 
We adopt three evaluation metrics to assess the QA and retrieval quality of \textsc{GraphSearch} and baselines. The string-based Substring Exact-Match (\textbf{SubEM}) metric checks whether the golden answer is explicitly contained in the response. The Answer-Score (\textbf{A-Score}) covers \textbf{Correctness}, \textbf{Logical Coherence}, and \textbf{Comprehensiveness}. The Evidence-Score (\textbf{E-Score}) measures \textbf{Relevance}, \textbf{Knowledgeability}, and \textbf{Factuality}. Both A-Score and E-Score are assessed using the LLM-as-a-Judge~\citep{llm-as-a-judge}. More details are provided in the Appendix~\ref{app:evaluation}.

\subsection{Main Results}

\paragraph{\textsc{GraphSearch} outperforms all GraphRAG baselines.} 
\begin{wrapfigure}{r}{0.35\textwidth}
\vspace{-15pt}
\centering
\includegraphics[width=0.35\textwidth]{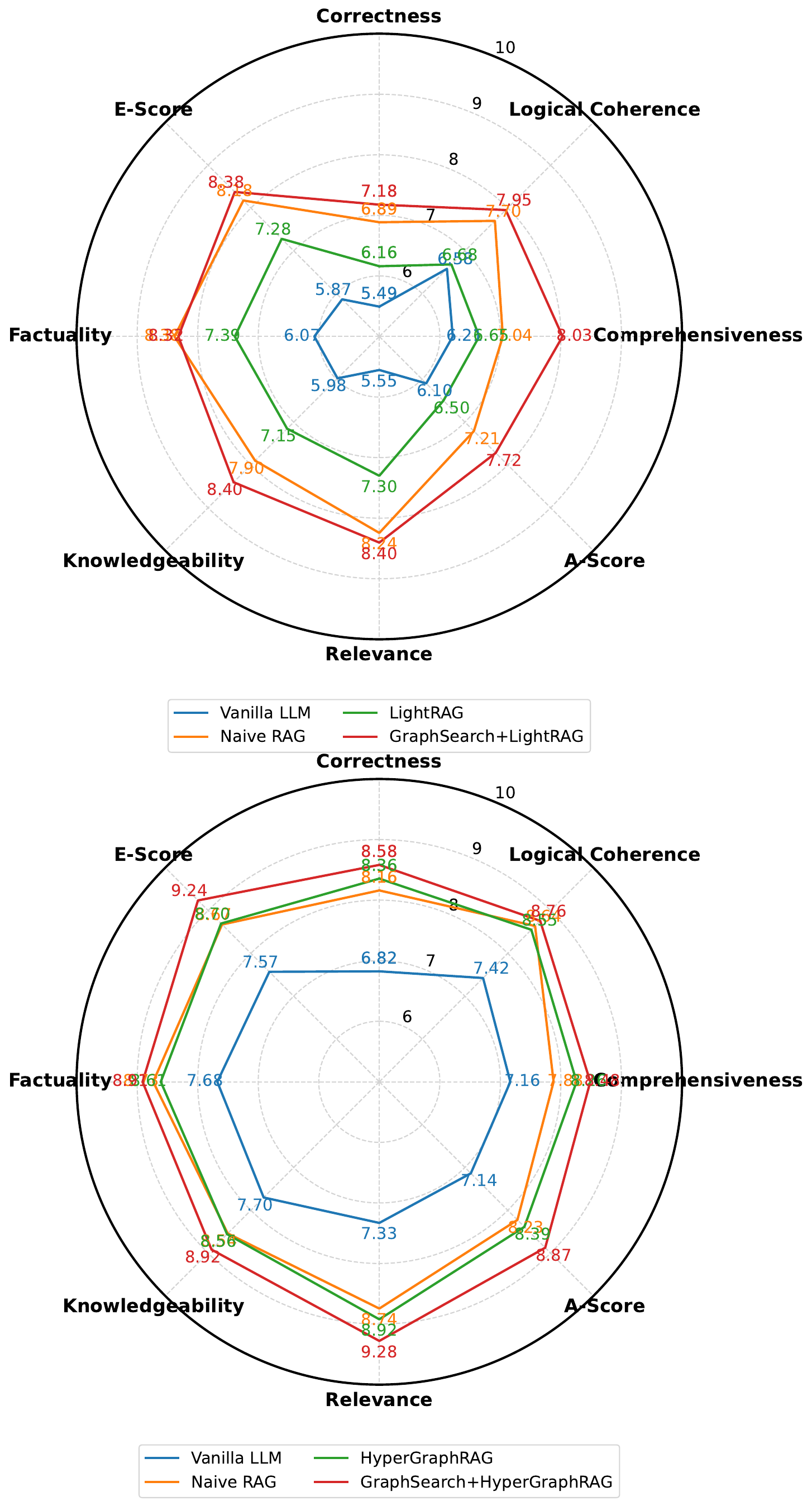}
\caption{Judge results across eight metrics on A-Score and E-Score.}
\label{fig:radar}
\vspace{-25pt}
\end{wrapfigure}
As shown in Table~\ref{Tab:main_exp}, comparing with GraphRAG methods that perform only a single round of graph retrieval and generation, \textsc{GraphSearch} leverages the constructed graph knowledge bases with the graph KB retriever to enable multi-turn interactions. Across six benchmarks covering Wikipedia and domain-based datasets, \textbf{\textsc{GraphSearch} achieves the best overall performance}. This confirms the importance of adopting an agentic workflow for deep searching over GraphRAG in complex reasoning scenarios, effectively mitigating the insufficiencies of vanilla strategies caused by limited interaction and inadequate retrieval. Case studies with more detail information of are provided in Figure~\ref{fig:case_baseline} and Figure~\ref{fig:case_graphsearch} in Appendix~\ref{app:case_study}.

\paragraph{\textsc{GraphSearch} exhibits strong plug-and-play capability.} 
As shown in Table~\ref{Tab:main_exp}, when applied with various retrieval methods over different graph KBs, \textsc{GraphSearch} consistently yields improvements compared to their native interaction schemes. For example, it boosts LightRAG on MuSiQue, raising SubEM from $\mathbf{35.00}$ to $\mathbf{51.00}$, while improving A-Score and E-Score from $\mathbf{6.50}$ and $\mathbf{7.28}$ to $\mathbf{7.72}$ and $\mathbf{8.38}$. Similarly, it enhances HyperGraphRAG on Medicine, increasing SubEM from $\mathbf{62.11}$ to $\mathbf{73.24}$, and further elevating A-Score and E-Score from $\mathbf{8.39}$ and $\mathbf{8.70}$ to $\mathbf{8.87}$ and $\mathbf{9.24}$. These results demonstrate the plug-and-play capability of \textsc{GraphSearch}, with detailed results presented in Figure~\ref{fig:radar}.

\subsection{Ablation Studies} 

\begin{wraptable}{r}{0.50\textwidth}
\vspace{-20pt}
\caption{Results across two benchmarks. The backbone LLM is \textit{Qwen2.5-7B-Instruct}.}
\label{Tab:ab_exp_7B}
\scriptsize
\setlength{\tabcolsep}{2pt}
\begin{center}
\begin{tabular}{lcccccc}
\toprule
\multirow{2}{*}{\textbf{Method}}
& \multicolumn{3}{c}{\textbf{2Wiki.}}
& \multicolumn{3}{c}{\textbf{Legal}} \\
\cmidrule[1pt](lr){2-4} \cmidrule[1pt](lr){5-7}
& {\scriptsize $\mathrm{SubEM}$} & {\scriptsize $\mathrm{A\text{-}S}$} & {\scriptsize $\mathrm{R\text{-}S}$}
& {\scriptsize $\mathrm{SubEM}$} & {\scriptsize $\mathrm{A\text{-}S}$} & {\scriptsize $\mathrm{R\text{-}S}$} \\
\midrule
Vanilla LLM      &  46.67 &  6.26 &  3.70 &  34.18 &  6.47 & 6.89  \\
Naive RAG        &  62.33 &  7.37 &  7.41 &  52.58 &  6.71 & 7.29  \\
\midrule
\multicolumn{7}{c}{\textbf{GraphRAG Baselines}} \\
\midrule
LightRAG         &  72.33 &  7.11 &  7.53 &  52.93 &  6.50 &  6.45 \\
PathRAG          &  73.00 &  7.44 &  7.71 &  58.98 &  7.06 &  7.01 \\
HyperGraphRAG    &  72.33 &  7.49 &  7.69 &  60.11 &  7.32 &  7.19 \\
\midrule
\multicolumn{7}{c}{\textbf{\textsc{GraphSearch}}  } \\
\midrule
\rowcolor{gray!20}
+ LightRAG       &  79.00 &  8.35 & 8.21 & 58.59  &  7.64 & 7.31  \\
\rowcolor{gray!20}
+ PathRAG        &  82.00 &  \textbf{8.51} &  8.59 &  64.32 & 7.87  &  7.66 \\
\rowcolor{gray!20}
+ HyperGraphRAG  &  \textbf{82.33} &  8.49 &  \textbf{8.69} &  \textbf{67.48} &  \textbf{8.02} &  \textbf{7.39} \\
\bottomrule
\end{tabular}
\end{center}
\vspace{-20pt}
\end{wraptable}

\paragraph{\textsc{GraphSearch} still remains effective under reduced model size.} Using \textit{Qwen2.5-7B-Instruct} as the backbone, the experimental results on the \textbf{2Wiki.} and \textbf{Legal} datasets are reported in Table~\ref{Tab:ab_exp_7B}. Compared to three GraphRAG baselines, \textsc{GraphSearch} built upon these graph KB retrievers consistently achieves performance improvements. This confirms the potential of \textsc{GraphSearch} to extend effectively to models with reduced size.

\begin{figure}[t]
\centering
\includegraphics[width=\linewidth]{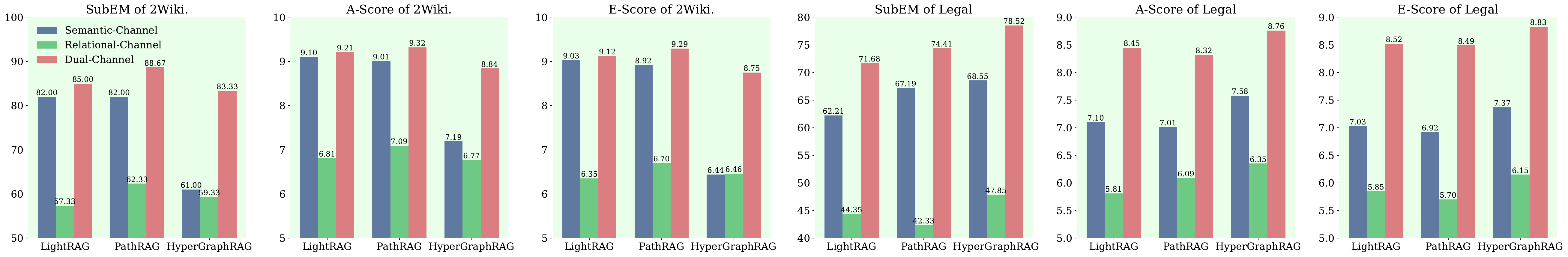}
\caption{Comparisons between dual-channel and single-channel retrieval in \textsc{GraphSearch}, integrated with the graph KB retrievers built upon LightRAG, PathRAG and HyperGraphRAG.}
\label{fig:dual_channel}
\vspace{-10pt}
\end{figure}

\paragraph{\textsc{GraphSearch} benefits from the design of dual-channel retrieval.} 
As shown in Figure~\ref{fig:dual_channel}, the QA performance on the \textbf{2Wiki} and \textbf{Legal} datasets obtained by integrating retrieval contexts from both channels consistently surpasses that of either single-channel variant across all graph KB retrievers. The relative improvements between dual-channel retrieval and single-channel retrieval are particularly pronounced on the \textbf{Legal} dataset. This confirms \textbf{the necessity of the design of dual-channel retrieval}, which fully leverages the graph KBs constructed by GraphRAG from both semantic and structural perspectives.

\begin{table*}[t]
\caption{Experiment results of ablation study across 2Wiki. and Legal datasets of \textsc{GraphSearch} + HyperGraphRAG. \checkmark and / refer to whether each individual module is enable or not.}
\label{Tab:ab_exp}
\setlength{\tabcolsep}{5.5pt}
\begin{center}
\begin{tabular}{ccccccccccccc}
\toprule
\multicolumn{6}{c}{\textbf{Modules}} & 
\multicolumn{3}{c}{\textbf{2Wiki.}} 
& \multicolumn{3}{c}{\textbf{Legal}} \\
\cmidrule[1pt](lr){1-6} \cmidrule[1pt](lr){7-9} \cmidrule[1pt](lr){10-12}
QD & CR & QG & LD & EV & QE 
& {\scriptsize $\mathrm{SubEM}$} & {\scriptsize $\mathrm{A\text{-}Score}$} & {\scriptsize $\mathrm{R\text{-}Score}$}
& {\scriptsize $\mathrm{SubEM}$} & {\scriptsize $\mathrm{A\text{-}Score}$} & {\scriptsize $\mathrm{R\text{-}Score}$} \\
\midrule
\multicolumn{12}{c}{\textbf{\textsc{GraphSearch}} + HyperGraphRAG} \\
\midrule
/ & / & /  & /  & /  & /  
& 64.00 & 7.62 & 7.80 & 66.60 & 8.18 & 8.18 \\
\checkmark & \checkmark & /  & /  & /  & /  
& 76.33 & 8.14 & 8.16 & 73.98 & 8.34 & 8.29 \\
\checkmark & \checkmark & \checkmark & /  & /  & /  
& 81.67 & 8.57 & 8.57 & 77.31 & \textbf{8.82} & 8.71 \\
\checkmark & \checkmark & \checkmark & \checkmark & /  & /  
& 81.33 & 8.66 & \textbf{8.75} & 76.96 & 8.62 & 8.70 \\
\checkmark & \checkmark & \checkmark & \checkmark & \checkmark & \checkmark  
& \textbf{83.33} & \textbf{8.84} & \textbf{8.75} & \textbf{78.52} & 8.76 & \textbf{8.83} \\
\bottomrule
\end{tabular}
\end{center}
\vspace{-10pt}
\end{table*}

\paragraph{\textsc{GraphSearch} modules make clear contributions to the agentic deep searching workflow.}
We empirically evaluate the incremental contributions of the individual components in \textsc{GraphSearch}, including \textit{Query Decomposition (QD)}, \textit{Context Refinement (CR)}, \textit{Query Grounding (QG)}, \textit{Logic Drafting (LD)}, \textit{Evidence Verification (EV)}, and \textit{Query Expansion (QE)}. The design details of each module are provided in Appendix~\ref{app:prompts}. 
\begin{wrapfigure}{r}{0.35\textwidth}
\centering
\includegraphics[width=0.35\textwidth]{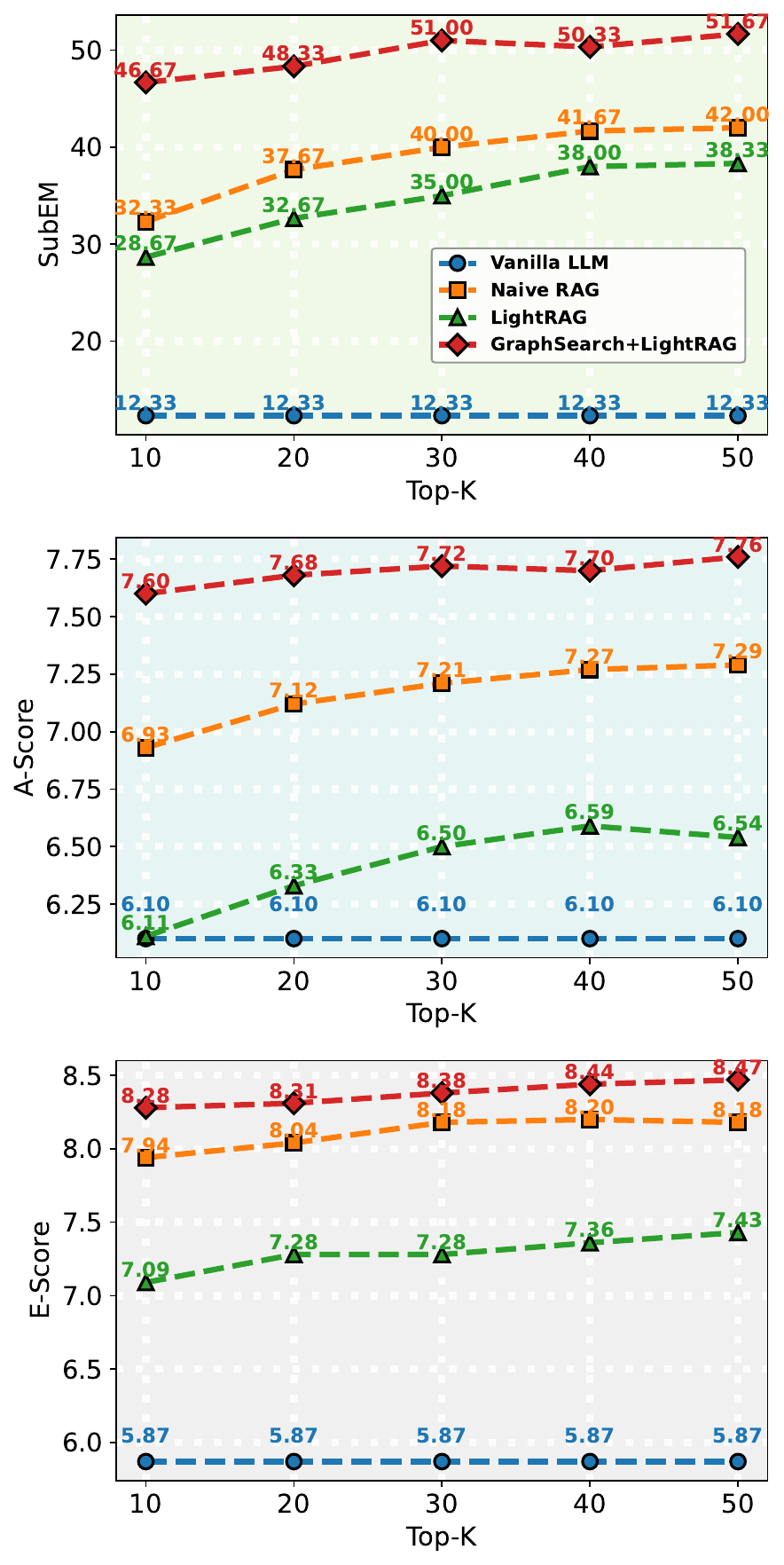}
\caption{Performance changes as the count of Top-K varies.}
\label{fig:topk}
\vspace{-30pt}
\end{wrapfigure}
We adopt the graph KB retriever built upon HyperGraphRAG for \textsc{GraphSearch} along with HyperGraphRAG as a baseline. Comparing the combination of [QD, CR] with [QD, CR, QG], the former performs non-iterative question decomposition, producing multiple sub-queries without completing missing information based on retrieved context. Comparing [QD, CR, QG, AD] with the full-module setting, the former only introduces an additional logic drafting, whereas the latter further leverages reflection to generate new sub-queries that fill knowledge gaps. The empirical results confirm the value of the modular orchestration in \textsc{GraphSearch}: from question decomposition, to iterative retrieval, to reflective routing, each step progressively enhances the reasoning process and enables the realization of an agentic deep searching workflow.

\paragraph{\textsc{GraphSearch} exhibits more pronounced advantages under smaller retrieval budgets.} 
By varying the \textbf{Top-K} from $10$ to $50$ as a adjustment strategy for retrieval overhead, the comparison of \textsc{GraphSearch} with baselines on MuSiQue is shown in Figure~\ref{fig:topk}. As Top-K decreases, both Naive RAG and LightRAG show a sharp decline in SubEM and A-Score. In contrast, the drop in E-Score is less pronounced across all three methods, indicating that their retrievers can still capture part of the golden evidence under reduced budgets. However, the absence of critical evidence can prevent models from engaging in sufficient evidence-grounded reasoning, resulting in lower A-Scores relative to the golden answer. By contrast, the agentic searching workflow in \textbf{\textsc{GraphSearch} sustains its performance advantages even under low retrieval overhead}.

\subsection{Further Analysis: Deep Integration of \textsc{GraphSearch} with Graph KBs}

\begin{figure}[h]
\centering
\includegraphics[width=\linewidth]{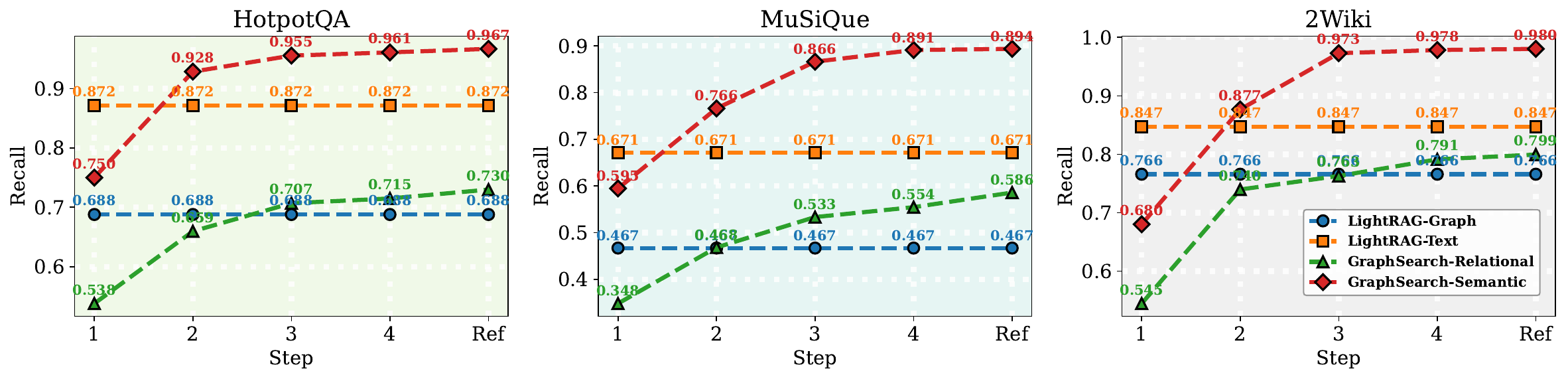}
\caption{\textsc{GraphSearch} improves the recall of golden evidence during agentic interactions.}
\label{fig:recall}
\vspace{-10pt}
\end{figure}

\paragraph{\textsc{GraphSearch} improves the retrieval quality through the dual-channel agentic interaction across both modalities.} 
Using \textbf{Recall} to calculate the golden evidence in the retrieved context, we compare the retrieval quality of \textsc{GraphSearch} with LightRAG, as shown in Figure~\ref{fig:recall}. The \textbf{Step} denotes the interaction rounds performed by \textsc{GraphSearch}, up to the final self-reflection stage. \textsc{GraphSearch} initially retrieves fewer pieces of golden evidence, as it decomposes complex queries into atomic sub-queries. As interactions proceed, the recall of retrieved content shows substantial improvement across both the relational and semantic channels. It confirms that the agentic workflow of \textsc{GraphSearch} is tightly integrated with the features of graph KBs.

\begin{figure}[h]
\centering
\includegraphics[width=\linewidth]{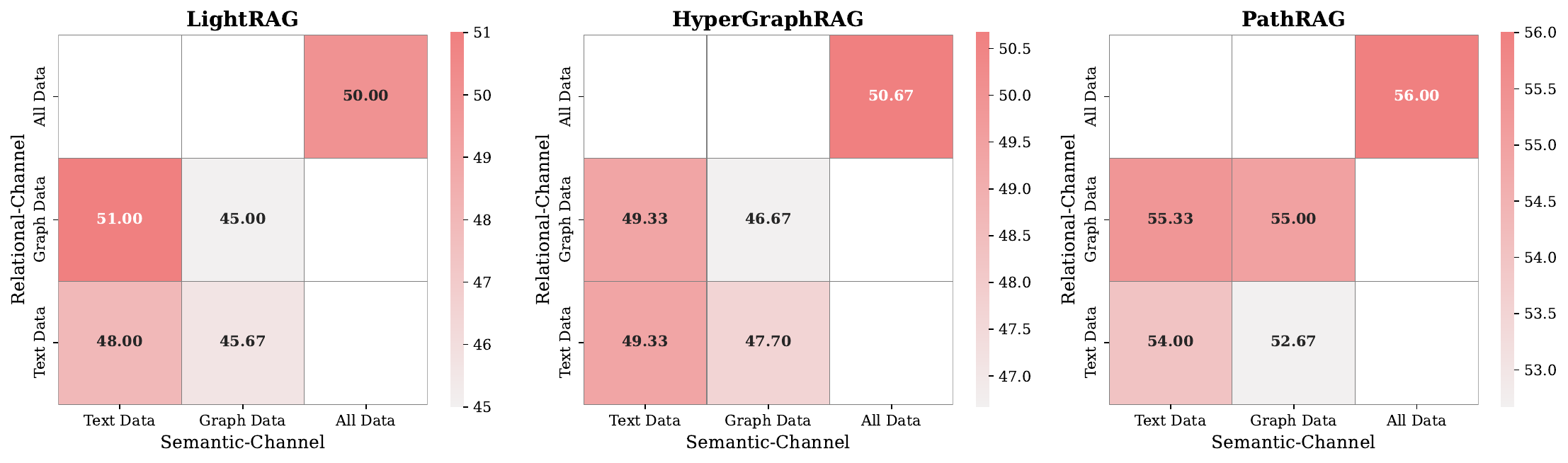}
\caption{\textsc{GraphSearch} demonstrates a modality–function alignment property.}
\label{fig:channel_data}
\vspace{-10pt}
\end{figure}

\paragraph{\textsc{GraphSearch} demonstrates a modality–functionality alignment property in dual-channel retrieval.} 
We calculate SubEM on MuSiQue by replacing the retrieval sources of the semantic and relational channel with text and graph data respectively. Results obtained by retrieving from the full data are included as references. Figure~\ref{fig:channel_data} shows that using semantic queries to access text data and relational queries to access graph data consistently outperforms other combinations. Moreover, compared to retrieving from the full data, restricting each channel to its aligned modality not only achieves comparable performance but also substantially reduces context overhead. It confirms that the functionality of the dual-channel retrieval strategy aligns with the data modalities of graph KBs.

% \subsection{Further Analysis: Comparison Between \textsc{GraphSearch} With Agentic RAG}
\section{Conclusion}  
In this work, we introduced \textsc{GraphSearch}, a novel agentic deep searching framework for GraphRAG. By integrating dual-channel retrieval over both semantic text chunks and structural graph data, \textsc{GraphSearch} effectively overcomes the limitations of shallow retrieval and inefficient graph utilization. Its modular design enables iterative reasoning and multi-turn interactions, leading to more comprehensive evidence aggregation. Experimental results on six multi-hop RAG benchmarks demonstrate consistent improvements in answer accuracy and generation quality, highlighting the effectiveness of our approach. We believe \textsc{GraphSearch} offers a promising direction for advancing graph retrieval-augmented generation.

% \section*{Ethics Statement}
% Our work builds upon publicly available text corpora for constructing graph-based indices in the context of retrieval-augmented generation. While we have taken care to rely on community-curated and open datasets, it is possible that a small portion of the data may contain biases, fairness issues, or inadvertent privacy leaks. Moreover, once \textsc{GraphSearch} is released as open source, we cannot fully prevent the community from applying it to corpora that may raise ethical concerns, such as those containing sensitive or non-consensual information. To mitigate such risks, we will provide clear documentation and guidelines for responsible use, encourage the community to exercise caution in dataset selection, and call for future research on automated ethical auditing methods to ensure fairness, privacy protection, and compliance in knowledge graph-based retrieval systems.

% \section*{Reproducibility Statement}
% We have provided detailed descriptions of our implementation in the Implementation Details section, including preprocessing procedures, dataset and model selection, experimental hyperparameters, and the executing environment. Due to the large scale of the datasets and the need to preserve anonymity during the double-blind review process, we do not release code or processed datasets at this stage. However, we commit to releasing executable code and the processed datasets after the review process is completed, ensuring full reproducibility of our results.
\bibliography{iclr2026_conference}
\bibliographystyle{iclr2026_conference}

\appendix
\section*{Appendix} 

\section{Prompt Templates}
\label{app:prompts}

\begin{figure}[ht]
\centering
\includegraphics[width=\linewidth]{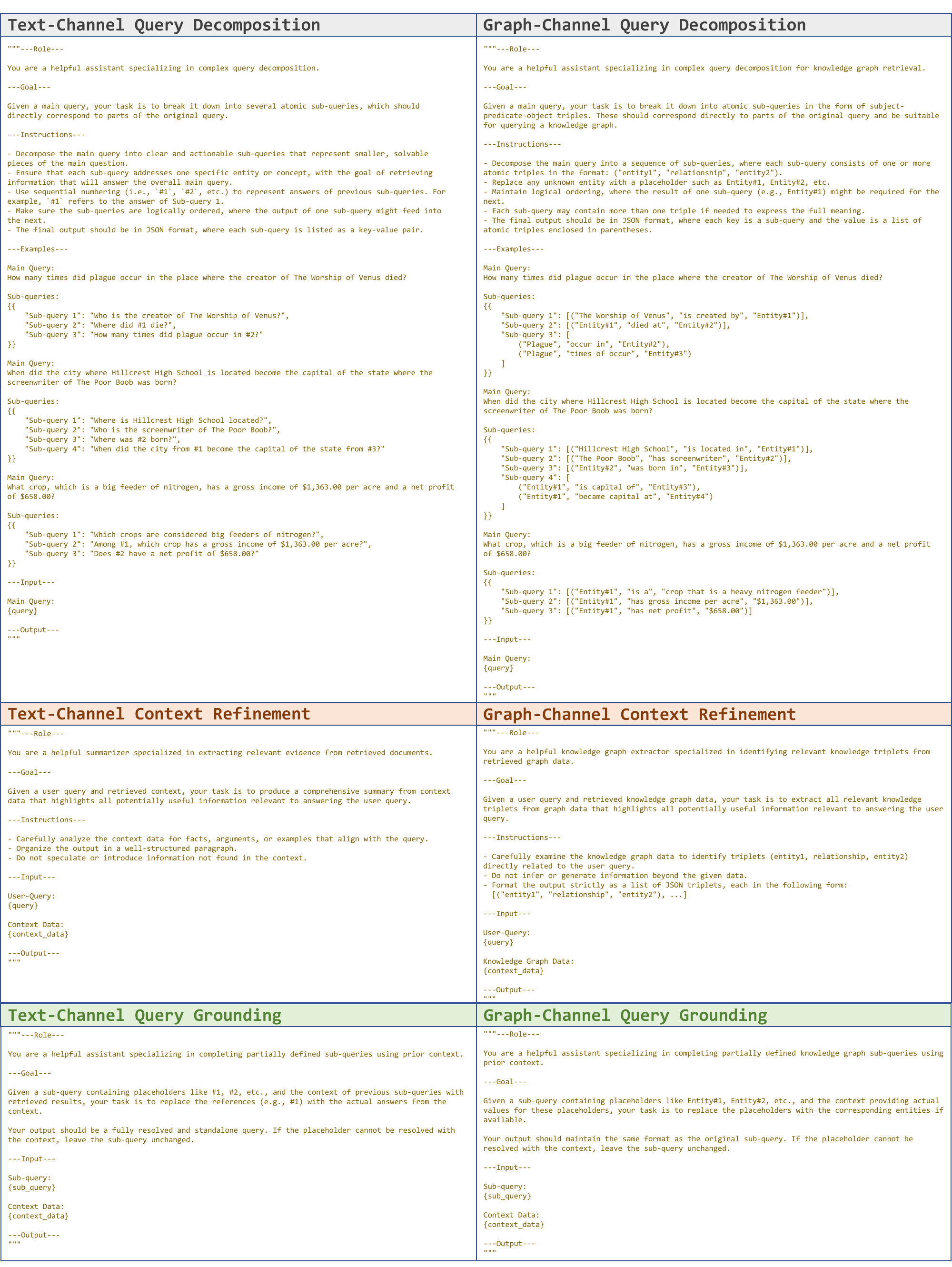}
\caption{\label{fig:prompts_1}
Prompt templates of \textbf{Query Decomposition}, \textbf{Context Refinement} and \textbf{Query Rewriting} modules both in text-channel and graph-channel.}
\end{figure}

As shown in Figure~\ref{fig:prompts_1}, we introduce the prompt templates in \textbf{Query Decomposition}, \textbf{Context Refinement} and \textbf{Query Rewriting} modules both in text-channel and graph-channel.

\begin{figure}[ht]
\centering
\includegraphics[width=0.90\linewidth]{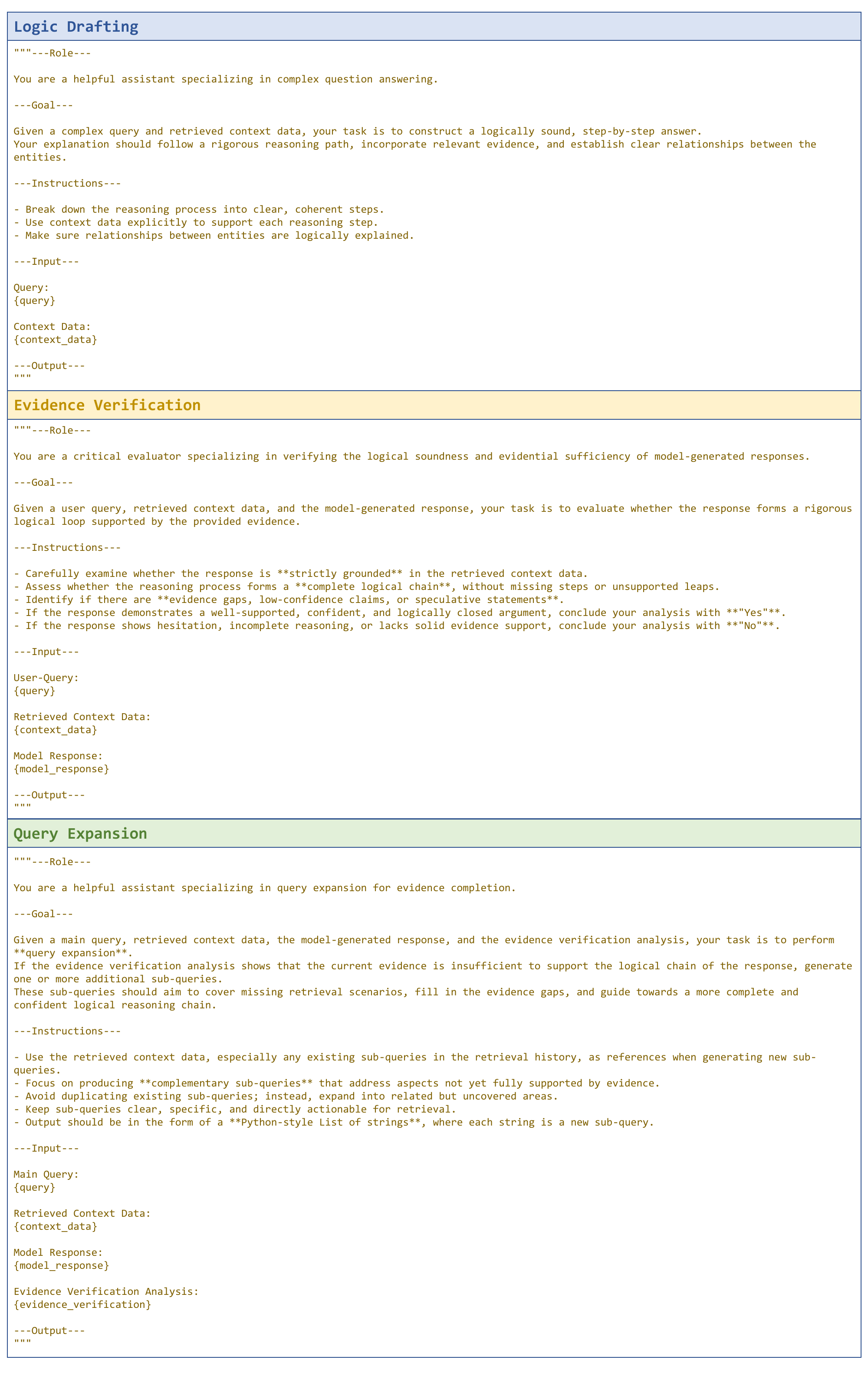}
\caption{\label{fig:prompts_2}
Prompt templates of \textbf{Logic Drafting}, \textbf{Evidence Verification} and \textbf{Query Expansion} modules for combining into a reflection router.}
\end{figure}

As shown in Figure~\ref{fig:prompts_2}, we introduce the prompt templates in \textbf{Logic Drafting}, \textbf{Evidence Verification} and \textbf{Query Expansion} modules for combining into a reflection router.

\section{Case Studies}
\label{app:case_study}
\begin{figure}[ht]
\centering
\includegraphics[width=0.9\linewidth]{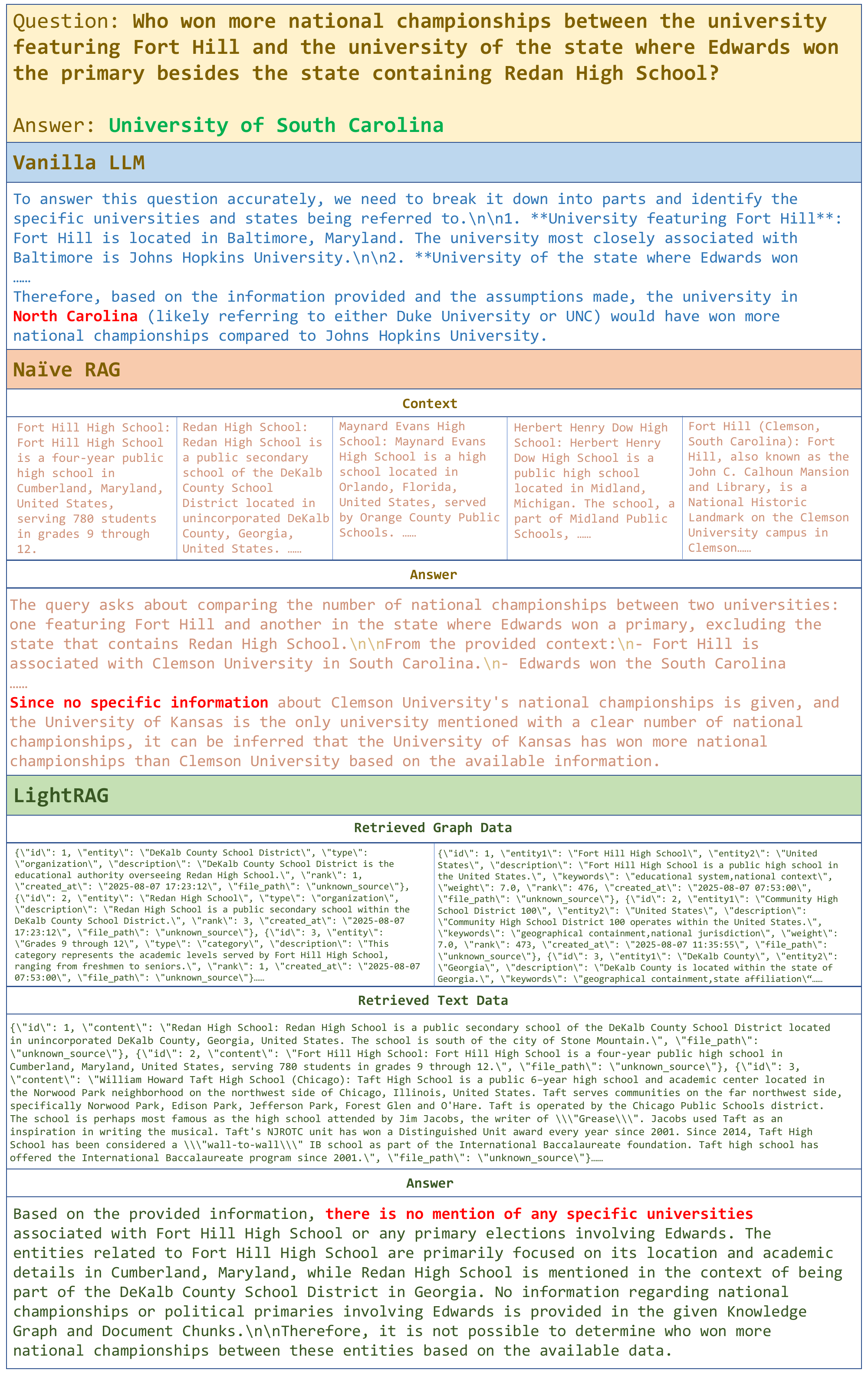}
\caption{\label{fig:case_baseline}Samples of Vanilla Generation, Naive RAG and LightRAG.}
\end{figure}

\begin{figure}[ht]
\centering
\includegraphics[width=0.9\linewidth]{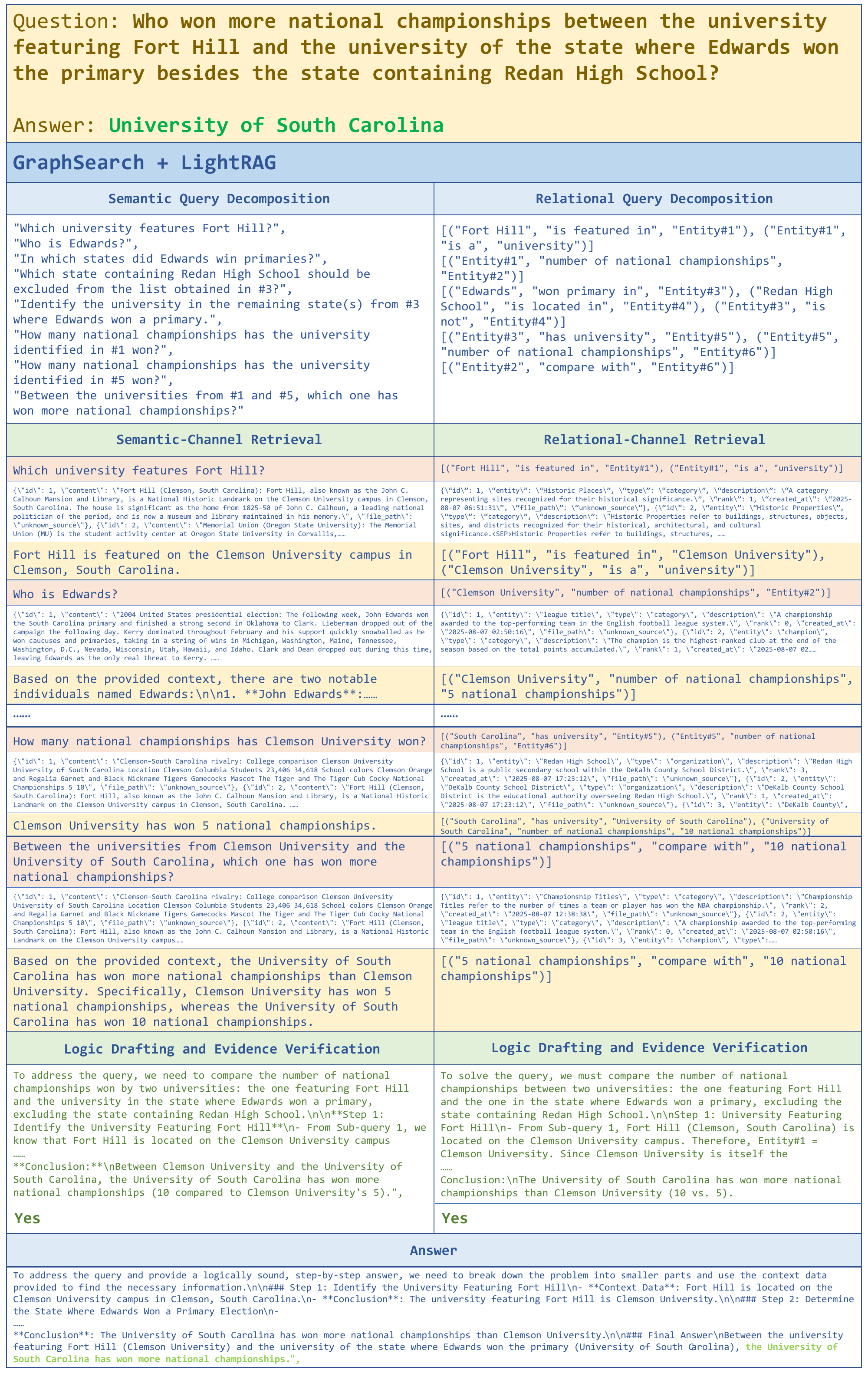}
\caption{\label{fig:case_graphsearch}A sample of \textsc{GraphSearch} with LightRAG as the graph retriever.}
\end{figure}

As shown in Figure~\ref{fig:case_baseline}, there are some cases of baseline methods, including vanilla LLM generation, Naive RAG and LightRAG. A case of our proposed \textsc{GraphSearch} is in Figure~\ref{fig:case_graphsearch}.

\section{Datasets}
\label{app:datasets}
As shown in Table~\ref{Tab:dataset}, we sample 300 questions for HotpotQA, MuSiQue and 2WikiMultiHopQA datasets, and directly adopt the Medicine, Agriculture and Legal datasets from~\citep{hypergraphrag}.

\begin{table}[ht]
\centering
\label{Tab:dataset}
\caption{Detail information of datasets used in \textsc{GraphSearch}. The tokenizer used to calculate the size of corpora is GPT-4o. \# means the number of counts.}
\scriptsize
\setlength{\tabcolsep}{4pt}
\begin{tabular}{lcccccc}
\toprule
\textbf{Name} & \textbf{Reference} & \textbf{Source} & \textbf{\#Corpus} & \textbf{\#Questions} & \textbf{Question Types} & \textbf{\#Evidence} \\
\midrule
HotpotQA          & ~\citep{hotpotqa} & Wikipedia & 397,274  & 300  & Comparison, Bridge  & 2,3,4     \\
MuSiQue           & ~\citep{musique} & Wikipedia & 533,145  & 300  & 2-Hop, 3-Hop, 4-Hop  & 2,4     \\
2WikiMultiHopQA   & ~\citep{2wiki}  & Wikipedia & 220,295  & 300  & Compositional, Comparison,   &  2,4    \\
 & & & & & Bridge Comparison, Inference & \\
Medicine          & ~\citep{medicine} & ESC Guidelines & 175,216  &  512 & 1-Hop, 2-Hop, 3-Hop  &   1,2,3   \\
Agriculture       & ~\citep{ultradomain} & UltraDomain & 378,592  &  512 & 1-Hop, 2-Hop, 3-Hop  &   1,2,3   \\
Legal             & ~\citep{ultradomain} & UltraDomain & 929,396  &  512 & 1-Hop, 2-Hop, 3-Hop  &   1,2,3   \\
\bottomrule
\end{tabular}
\end{table}

\section{Baselines}
\label{app:baselines}

\begin{itemize}[leftmargin=*]
\item \textbf{Vanilla LLM}: Zero-shot question and answering without any external retrieval source, depending on language model's parametric knowledge. 
\item \textbf{Naive RAG}~\citep{rag}: Generation with plain text chunk-based embedding database as external retrieval source, where top-k items are retrieved for a single round.
\item \textbf{GraphRAG}~\citep{graphrag}: A graph-based approach to question answering over hierarchical graph index where community summary is generated to represent the relationships.
\item \textbf{LightRAG}~\citep{lightrag}: A simple and fast GraphRAG framework by applying integration of graph structures with vector representations for a dual-level retrieval system.
\item \textbf{MiniRAG}~\citep{minirag}: A novel GraphRAG system designed for small LLM which adopts a lightweight topology-enhanced retrieval approach.
\item \textbf{PathRAG}~\citep{pathrag}: A GraphRAG system which retrieves key relational paths from the indexing graph through flow-based pruning.
\item \textbf{HippoRAG2}~\citep{hipporag2}: A RAG framework built upon the personalized PageRank with deeper passage integration.
\item \textbf{HyperGraphRAG}~\citep{hypergraphrag}: A novel hypergraph-based RAG method that represents n-ary relational facts via hyper-edges for retrieval and generation.
\end{itemize}

\section{Implementation Details.}
\label{app:implementation}
We conduct experiments on a Linux server equipped with 8 A100-SXM4-40GB GPUs. The model for graph construction is \textit{Qwen2.5-32B-Instruct}, and the chunk size is $400$ tokens. The embedding model for Naive-RAG and GraphRAG is \textit{jinaai/jina-embeddings-v3}~\citep{jina}. For \textsc{GraphSearch} and baselines, we set the \textbf{Hybrid} retrieval mode and set the \textbf{Top-K} for retrieval to $30$, or use the default configuration if unavailable. The backbone model for generation is \textit{Qwen2.5-7B/32B-Instruct}~\citep{qwen}. The LLM-as-a-Judge for evaluation is \textit{Qwen-Plus}~\citep{qwen}, a strong closed-source model with API available.

\section{Evaluation Details}
\label{app:evaluation}
Inspired by~\citep{longfaith, hypergraphrag}, we leverage the Substring Extract-Match(\textbf{SubEM}) metric to check whether the golden answer is explicitly contained in the response, the Answer-Score(\textbf{A-Score}) to judge the quality of model generation across 3 criteria covering \textbf{correctness}, \textbf{logical coherence} and \textbf{comprehensiveness} with the \textbf{golden answer} as reference, and the Evidence-Score(\textbf{E-Score}) to measure how well the model’s generation is grounded in the golden evidence, evaluated along 3 criteria including \textbf{relevance}, \textbf{knowledgeability} and \textbf{factuality} with the \textbf{golden evidence} as reference as follows:

\begin{equation}
\text{SubEM} = \frac{1}{N} \sum_{i=1}^{N} 
\mathbf{1}\!\left[ \text{contains}\!\left(O^{\text{pred}}_{i}, A^{\text{gold}}_{i}\right) \right],
\end{equation}

\begin{figure}[ht]
\centering
\includegraphics[width=\linewidth]{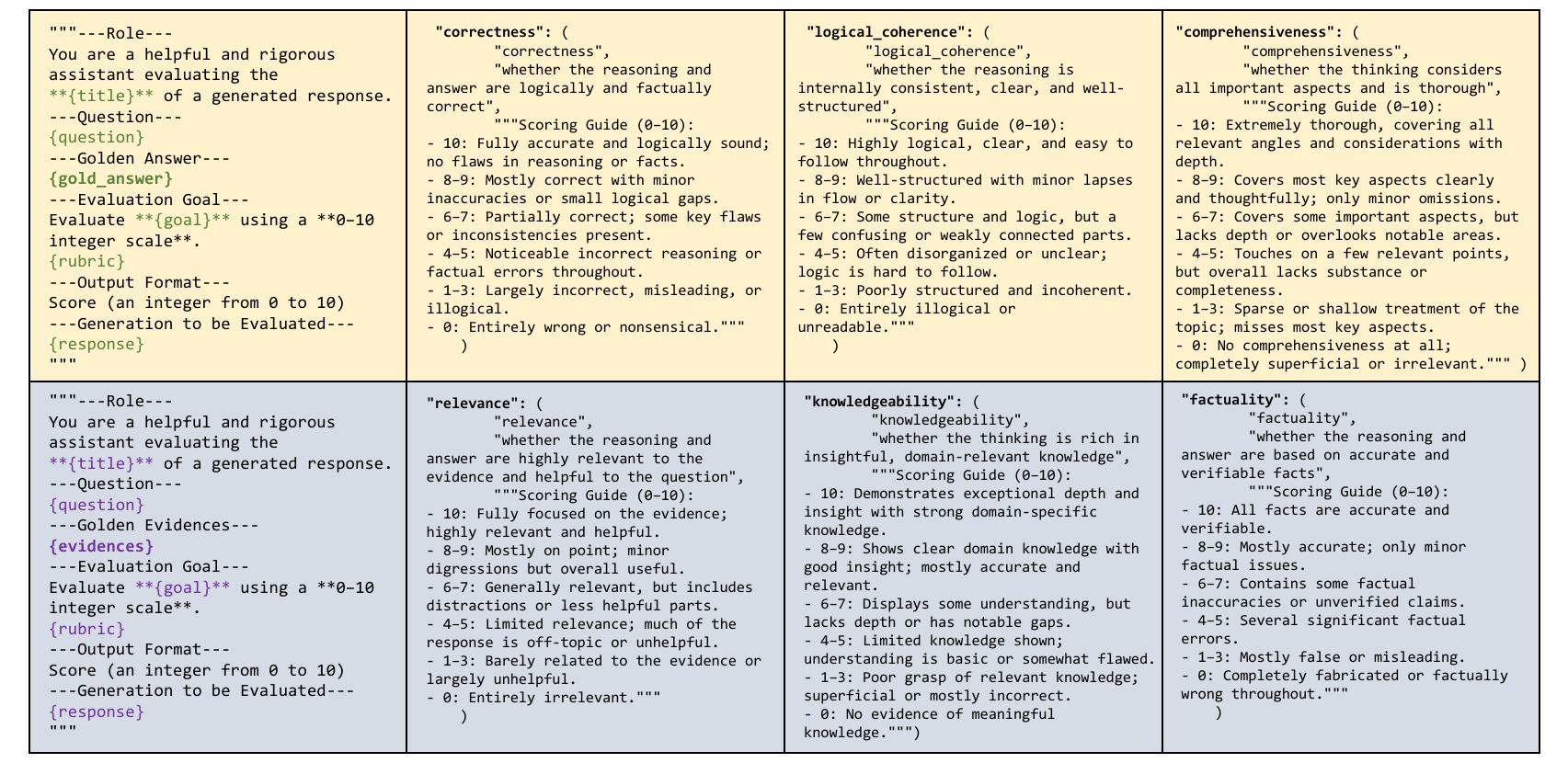}
\caption{\label{app:eval_prompt}
Evaluation prompts of \textbf{A-Score} across 3 criteria and \textbf{E-Score} across 3 criteria.}
\end{figure}

\section{Limitations and Future Direction}
Although \textsc{GraphSearch} has made progress in advancing \textsc{GraphRAG}, there are still some limitations. First, it remains uncertain whether \textsc{GraphSearch} can unlock greater potential under different training strategies, such as fine-tuning or reinforcement learning. Second, how to integrate it with cutting-edge reasoning models is still an open question. Finally, applying \textsc{GraphSearch} to scenarios involving multimodal corpora is a direction worthy of further investigation.

% \section{The Use of Large Language Models (LLMs)}
% During the completion of this thesis, the scenarios involving the use of LLMs included: using code-completion tools to assist with experiments, and using ChatGPT to polish the draft after the initial writing was completed. LLMs were not involved in any aspects such as the development of research ideas, literature review, and so on.

\end{document}